\documentclass[10pt,twocolumn,letterpaper]{article}

\usepackage{iccv}
\usepackage{times}
\usepackage{epsfig}
\usepackage{graphicx}
\usepackage{amsmath}
\usepackage{amssymb}
\usepackage[accsupp]{axessibility} 
\usepackage{booktabs}
\usepackage{mathtools}
\newcommand{\tabincell}[2]{\begin{tabular}{@{}#1@{}}#2\end{tabular}}

\newcommand{\ttsize}{\footnotesize}
\usepackage[pagebackref=true,breaklinks=true,letterpaper=true,colorlinks,bookmarks=false]{hyperref}

\iccvfinalcopy 

\def\iccvPaperID{7253} 

\ificcvfinal\fi

\begin{document}

\title{GroupFormer: Group Activity Recognition with Clustered Spatial-Temporal Transformer}

\author{Shuaicheng Li$^{1*}$, Qianggang Cao$^{1*}$, Lingbo Liu$^{2}$, Kunlin Yang$^{1\dagger}$, Shinan Liu$^{1}$, Jun Hou$^{1}$, Shuai Yi$^{1}$\\
$^{1}$Sensetime Research, $^2$The Hong Kong Polytechnic University
\\
{\tt\small \{lishuaicheng,caoqianggang,yangkunlin,liushinan,houjun,yishuai\}@sensetime.com}\\
{\tt \small liulingbo918@gmail.com}
}

\maketitle
\ificcvfinal\thispagestyle{empty}\fi

\begin{abstract}
%
Group activity recognition is a crucial yet challenging problem, whose core lies in fully exploring spatial-temporal interactions among individuals and generating reasonable group representations. However, previous methods either model spatial and temporal information separately, or directly aggregate individual features to form group features. To address these issues, we propose a novel group activity recognition network termed GroupFormer. 
It captures spatial-temporal contextual information jointly to augment the individual and group representations effectively with a clustered spatial-temporal transformer. 
%
Specifically, our GroupFormer has three appealing advantages: 
(1) A tailor-modified Transformer, Clustered Spatial-Temporal Transformer, is proposed to enhance the individual representation and group representation.
(2) It models the spatial and temporal dependencies integrally and utilizes decoders to build the bridge between the spatial and temporal information.
(3) A clustered attention mechanism is utilized to dynamically divide individuals into multiple clusters for better learning activity-aware semantic representations.
Moreover, experimental results show that the proposed framework outperforms state-of-the-art methods on the Volleyball dataset and Collective Activity dataset.
Code is available at \url{https://github.com/xueyee/GroupFormer}
\end{abstract}
\section{Introduction}
\let\thefootnote\relax\footnotetext{$*$ \textrm{\ indicates\ equal\ contribution.}}
\let\thefootnote\relax\footnotetext{$\dagger$ \textrm{\ denotes\ corresponding\ author.}}

Group activity recognition is a critical studied problem due to its wide applications in surveillance systems, video analysis, and social behaviors analysis.  
Different from conventional action recognition, group activity recognition concentrates on understanding the scene of multiple individuals. 
The intuitive tactic to recognize group activity is to model relevant relations between individuals and infer their collective activity.
Nevertheless, exploiting individual relations for inferring collective activity is very challenging, especially due to the complicated variations in both spatial and temporal transition in untrimmed scenarios.

\begin{figure}[t]
\centering
\includegraphics[width=1.0\linewidth]{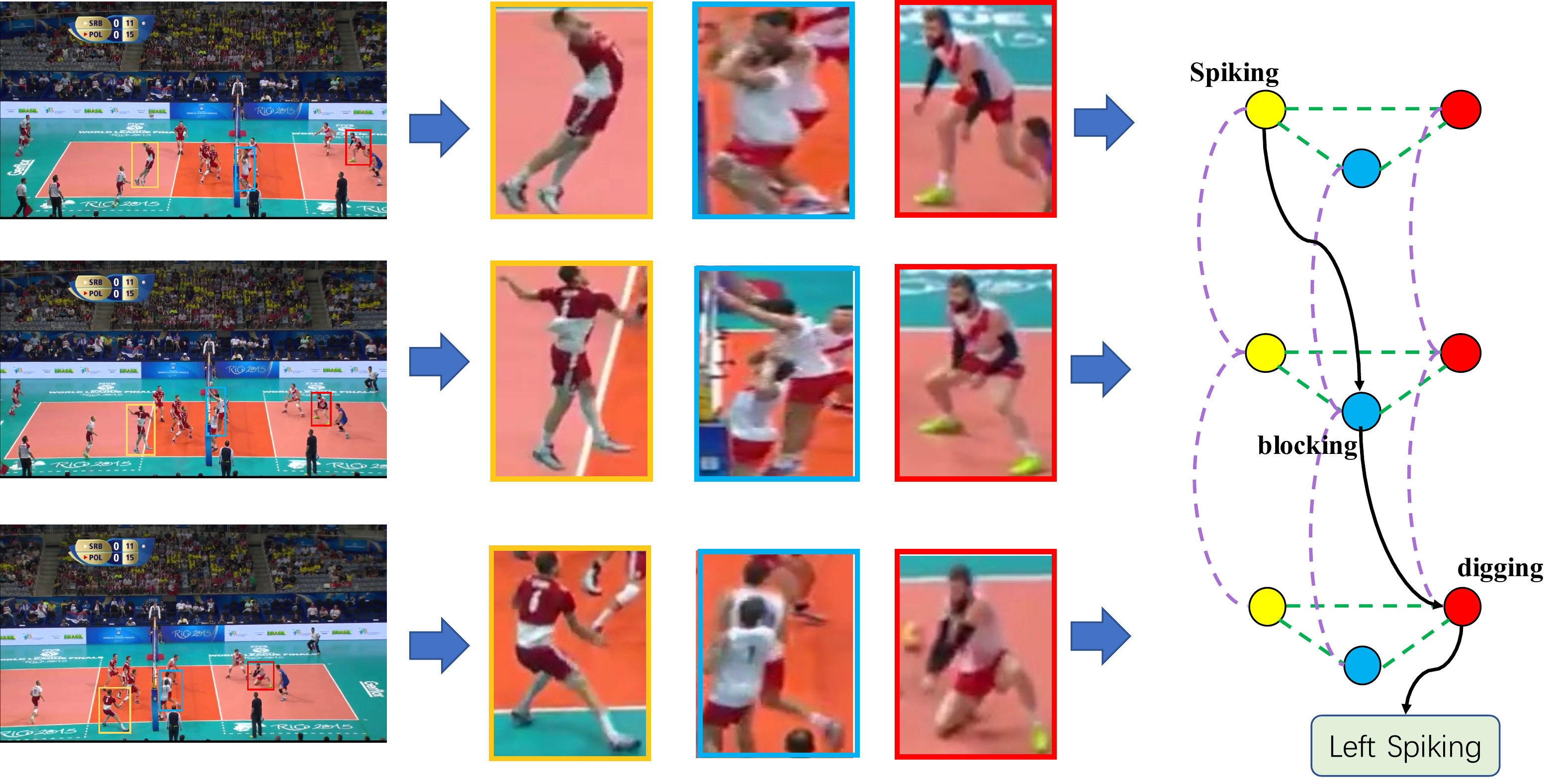}
\caption{Examples of a clip centered at the annotated frame. The actors with `spiking', `blocking', `digging' actions perform temporally, but they may perform strong spatial-temporal dependencies, which shows the importance of considering spatial and temporal interactions for reasoning about the `Left Spiking' activity.}
\label{fig:intro}
\end{figure}

Sorts of efforts have been dedicated to capturing relation context in videos for the sake of group activities inferring.
Earlier methods \cite{Social-scene, hierarchical, SBGAR,CERN, stagnet, structure2016} utilized recurrent neural network (RNN) to model the dynamics of the individuals, which require a large amount of computational cost.
More recent works \cite{actorTR2020, ARG2019,joint2020,empowering2020,progressive2020} applied attention-based methods to model the individual relations for inferring group activity. 
\cite{ARG2019} built relational graphs and considered the actor interactions in several frames.
\cite{empowering2020} captured spatial and temporal self-attention respectively, which are added and utilized to reinforce the mean-field CRF \cite{CRF}.
\cite{actorTR2020} introduced the standard Transformer encoder as the feature extractor to selectively exploit the spatial actor relations without considering the temporal dynamical information.
\cite{progressive2020} proposed a relation learning network to model and distill the group-relevant actions and activities by two agents.

However, the aforementioned methods confront two challenges that remain to be addressed: 1) build a bridge to model spatial-temporal contextual information integrally and 2) group individuals based on their inter-connected relations for better inferring the global activity context. 
In the former case, few of the previous methods completely consider the spatial and temporal dependencies in a joint model, while diverse time-series information has strong spatial dependencies, as illustrated in Figure \ref{fig:intro}. Therefore, capturing spatial-temporal dependencies jointly is critical for reasoning about the group activity.  
In the latter case, full-connected relations introduced by prior methods \cite{ARG2019, actorTR2020} is suboptimal since interference information of irrelevant individuals is introduced.
Intuitively, not all individuals' relations in the multi-person scenario perform key impacts for the inferring of group activity.
As shown in Figure \ref{fig:intro}, in the volleyball scenario, the interactions between actors with `spiking' and `blocking' are considerably higher than the relations between actors with `spiking' and  `standing', which contribute more for group activity inferring. 
In other words, the group activity is usually determined by a critical group of individuals with underlying closet relations.

In this paper, we propose an end-to-end trainable framework termed GroupFormer, which utilizes a tailor-modified Transformer to model individual and group representation for group activity recognition.
Firstly, we develop a Group Representation Generator to generate an initial group representation by merging the individual context and scene-wide context. 
Multiple stacked Spatial-Temporal Transformers (STT) are then deployed to augment and refine both the individual and group representation. 
Specifically, we adopt encoders to embed the spatial and temporal features and apply decoders in a cross manner to build a bridge to model the spatial and temporal contextual information integrally.
And a decoder is employed to contextualize the individual representation for augmenting group representation.
Besides, in contrast to the existing full-attention manner in Transformer, our STT is further enhanced by a clustered attention mechanism, referred to as Clustered Spatial-Temporal Transformer (CSTT), to model inter-group relations and intra-group relations.
More specifically, we dynamically divide all individuals into $C$ clusters, where individuals in the same cluster usually have relevant semantic information.
By performing information propagation within each cluster, we can generate compact action features of individuals. 
Our inter-group attention is to fully model the relations among clusters for facilitating the group activity-aware representation learning.
Finally, experimental results show that the proposed network outperforms state-of-the-art methods on the widely adopted Volleyball and Collective Activity datasets.
In short, the contributions of this work can be summarized as three folds: 
\begin{itemize}
\item
We propose a new group activity recognition framework, termed GroupFormer, which takes advantage of query-key mechanism to model spatial-temporal context jointly for group activity inferring.
\item
A clustered attention mechanism is introduced to assign individuals into groups and build inter- and intra-group relations to enrich the global activity context.
\item
We perform extensive experiments on the widely adopted Volleyball and Collective datasets.
The results show that our GroupFormer outperforms the state-of-the-art methods by a significant margin.
\end{itemize}

\begin{figure*}[t]
\centering
\includegraphics[width=0.8\linewidth]{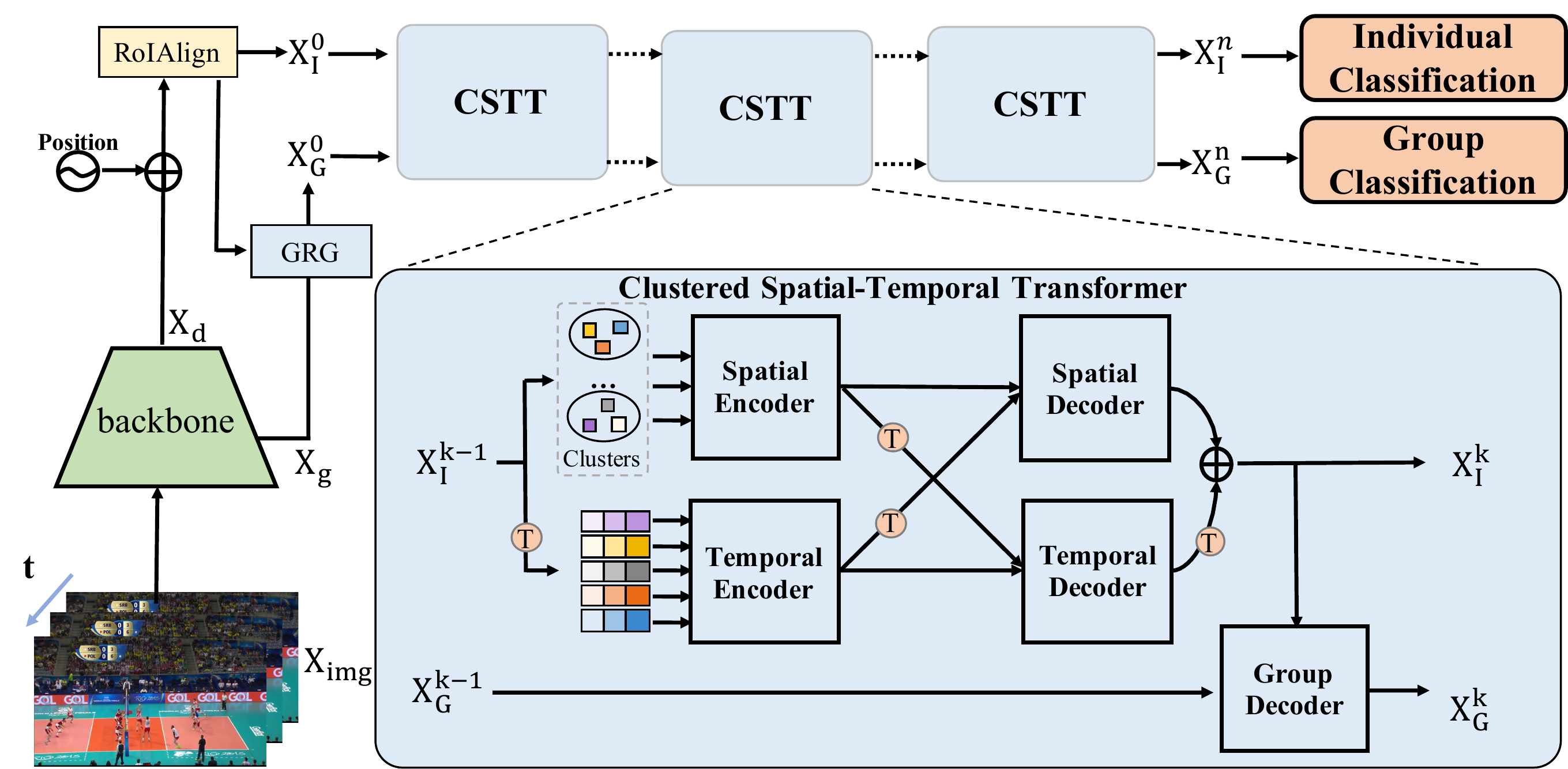}
\caption{Illustration of our proposed GroupFomer. It contains three main components: 1) a CNN backbone that extracts feature representation of video clips. 2) a Group Representation Generator that initializes the group representation from individual and scene features. 3) a Clustered Spatial-Temporal Transformer that models the spatial-temporal relations and refines the group representation and individual representation.}
\label{fig:overview}
\end{figure*}

\section{Related Works}

\subsection{Group Activity Recognition}
Group activity recognition has been extensively received more attention due to its wide applications.
Traditional approaches typically extracted hand-crafted features and then processed by probabilistic graphical models \cite{sum2015,monte2013,hirf2014,discriminative2011,lan-social2012,bilinear2013} and AND-OR grammar methods   \cite{cost2012,shu-joint2015}.
With the rapid development of deep learning, RNN-based methods have achieve remarkable performance due to the learning of temporal context and high-level information \cite{Social-scene, structure2016, hier2018, hierarchical, SBGAR, shu2019hierarchical, CERN}. 
\cite{hierarchical} proposed a LSTM model to capture the temporal evolution of each individual and then generate the holistic representation by pooling the actor-leave features.
\cite{CERN} deployed a two-level hierarchy of LSTMs to recognize the group activities more reliably by minimizing the energy of the predictions.
In \cite{hier2018}, a relational layer is introduced to capture spatial relations for each person for group representation generating.
\cite{stagnet} designed a semantic graph and extended it to the temporal dimension by RNN to integrate the spatial and temporal relations.

In recent works, \cite{ARG2019} introduced the graph model and built an actor relation graph using GCN to augment the individuals' representation, while the constructed relational graphs are limited to few frames and don't consider the latent temporal individual relationships.
\cite{joint2020} adopted self-attention mechanism and graph attention module to model spatial individual dependencies and utilized I3D backbone to capture temporal context.
It only captured temporal information using 3D-CNN based backbone and didn't model the temporal relations in their own module.
\cite{empowering2020} made use of temporal and spatial relations with stacked manner to reinforce the mean-field CRF for learning spatial relations and temporal evolution of the actors. 
The most related work is \cite{actorTR2020}, which also introduced the transformer to capture semantic representation.
However, it only utilized the encoder of vanilla Transformer as a feature extractor for modeling spatial dependencies, and paid much more attention to build strong activity representations using several branches backbone without integrally considering spatial and temporal dependencies.
To better exploit the spatial-temporal dependencies integrally, we propose CSTT to not only utilize the encoder to augment semantic representation, but also use a decoder to build a bridge between spatial and temporal relations.

\subsection{Transformer}
Transformer was first proposed in \cite{attention2017} for sequence-to-sequence machine translation task, and since then has been widely adopted in various natural language processing tasks.
The adopted self attention mechanism is particularly suitable for capturing long-term dependencies.
Based on Transformer, a series of modifications \cite{bert2018,SparseTR2019,efficient2021} have been designed to address the limitations in the standard transformer such as  computation bottleneck.\\
\textbf{Transformer in Vision.}
The attention mechanism has been widely used in computer vision domain.
Due to quadratic cost for the naive application of transformer to images, each pixel attends to all the pixels with the query-key mechanism.
Therefore, earlier methods \cite{nonlocal2018,attention2019} only utilized the self-attention to capture long-range context.
\cite{nonlocal2018} designed a non-local attention module to capture long-term dependencies in computer vision tasks.
\cite{attention2019} employed a 2D self-attention mechanism to selectively replace 2D convolution and achieve better results than original 2D convolution layer.
Very recently, Vision Transformer (ViT) \cite{vit2020}, dividing an image into $16 \times 16$ patches, took these patches as input and fed them into a standard transformer.
The simplicity incurs an extremely high computational price.
DETR \cite{DETR2020} largely streamlined the detection pipeline based on Transformers, and achieved stronger performances compared with previous CNN based detectors.
Unlike ViT \cite{vit2020}, DETR adopted CNN based backbone for extracting low-level features and encoder-decoder based transformer for exploiting high-level concepts.\\
\textbf{Spatial-Temporal Transformer.}
While it has not been explicitly stated, using attention mechanism to model spatial-temporal contexts is quiet general and therefore encompasses many previous works.
\cite{spatial2020} designed a variant of graph neural network to model the time-varying spatial dependencies.
Attention-based spatial-temporal GCN proposed by \cite{attentiontf-2019} captured dynamic correlation from the message passing of the graph to learn spatial and temporal features.
\cite{nast2021} proposed to employ transformer-based encoders to build spatial and temporal attentions separately and index the temporal attention to the corresponding spatial attention matrix directly.
\cite{space2021} simply extended the ViT \cite{vit2020} design to video by proposing several scalable schemes for space-time self-attention.
Previous methods either capture spatial and temporal contexts in stacked manners \cite{space2021,spatial2020}, or utilize parallel modules to extract spatial and temporal features and then fuse them simply \cite{nast2021}.
Different from them, we embed spatial and temporal context in parallel, and adopt a decoder to complementary exploit spatial-temporal contextual information.

\section{GroupFormer}
The proposed GroupFormer is tailored for the group activity recognition involving multiple individuals.
The overview of GroupFormer is illustrated in Figure \ref{fig:overview}.
We first process the input image using CNN backbone in Section \ref{sec:extractor} and then initialize the group representation in Section \ref{sec:GRG}.
Next, in Section \ref{sec:ST} we introduce our Spatial-Temporal Transformer and clustered attention mechanism in detail.
\subsection{Feature Extractor} \label{sec:extractor}
We adopt Kinetics \cite{Kay:2017:Kinetics} pretrained inflated 3D network (I3D) \cite{2017:I3D} as the backbone, and take RGB video clips as input.
We start by slicing out a $T$-frames centered the annotated frame, which is denoted as  $\mathbf{X_{img}} \in \mathbb{R}^{T \times 3 \times H \times W }$ (with 3 color channels).
In practice, we extract feature maps from the last convolutional layer and resize to $\mathbf{X_g} \in \mathbb{R}^{T \times C_g \times H' \times W'}$, which can be viewed as the scene feature for the entire video clip.
We also generate the higher resolution feature map $\mathbf{X_d}$ from the intermediate \textit{Mixed\_4f} layer, and a learnable positional encoding mentioned in \cite{attention2017,vit2020} is added to the feature map $\mathbf{X_d}$.
RoIAlign \cite{mask2017} is applied to extract features for each individuals given $N$ bounding boxes in each video frame.
In addition, the pose information of each actor is obtained by AlphaPose \cite{alphapose2017} and concatenated with the above individual features to provide the final individual features.
A fully connected layer is finally adopted to embed the aligned individual features into a $D$ dimensional feature vector for each actor, which can be packed into together named individual representation $\mathbf{X_I} \in \mathbb{R}^{T \times N \times D}$.
Besides, we have also conducted experiments on Inception-v3 \cite{inception2016} backbone followed the implementation \cite{ARG2019} for a fair comparison with previous methods.

\subsection{Group Representation Generator}\label{sec:GRG}
Group Representation Generator (GRG) is a preprocessed component for initializing group representation in our model.
Inspired by \cite{nonlocal2018,latentgnn2019,visual2020}, video frames can be summarized by a set of feature vectors called visual tokens. Therefore, we transform the scene features and individual features into several tokens respectively and then aggregate them to generate group representation.
For scene feature $\mathbf{X}_g$, we view the time dimension as batch dimension and apply a 2D convolution to summarize $C_g$ channels into $K$.
We reshape it into a flattened features $\mathbf{X'}_g \in \mathbb{R}^{T \times K \times (H' \cdot W')}$ and implement a $\mathrm{softmax}$ operation to generate the spatial attention matrix $\mathbf{A} \in \mathbb{R}^{T \times K \times (H' \cdot W')}$.
Afterwards, we adopt a 2D convolution to embed the scene features into $D$ channels and compute the weighted sum of every pixels with $\mathbf{A}$ to generate $K$ visual scene tokens followed by an $\mathrm{Avgpool}$ operation to achieve a scene token shaped as $T \times D$ (we set $K$ to $8$).
For aligned individual features $\mathbf{X_I^0}$, we feed a learned query shaped as $T \times D$ and individual features into a decoder to generate an individual token.
In the end, we fuse the individual token and scene token to form a group token termed initialized group representation $\mathbf{X_G^0} \in \mathbb{R}^{T \times D}$.
\subsection{Clustered Spatial-Temporal Transformer} \label{sec:ST}
\subsubsection{Canonical Transformer} \label{sec:CT}

In this section, we briefly revisit the standard Transformer architecture.
The canonical Transformer \cite{attention2017} is proposed for the sequence-to-sequence task such as language translation.
It contains encoders and decoders, both of which apply stacked multi-head attention layers and feed-forward network.
The multi-head attention computes the weights by comparing the pairwise similarity between the one and all others, which shows remarkable performance in capturing long-term dependencies.
In brief, a sequence of $l$ vectors with the dimension $d$, represented as $f \in \mathbb{R}^{l \times d}$, is first mapped to the \textit{query}, \textit{key} and \textit{value} using linear projections respectively.
Attention matrix is obtained from the scaled dot product of the \textit{query} and \textit{key}.
The output features are computed as a weighted sum of the \textit{values} based on the attention weights.
Furthermore, The feed-forward network (FFN), consisting of linear transformation and non-linear activation functions, is used to embed the features.
More detailed description of the original architecture can be referred to literature \cite{attention2017,imageTR2018}.

\subsubsection{Spatial-Temporal Transformer} \label{sec:STTR}
Our Spatial-Temporal Transformer (STT) tailored for group activity recognition is designed carefully for augmenting individual representation and group representation.
It includes two encoders (a spatial encoder and temporal encoder) in parallel to generate spatial and temporal features respectively.
The individual decoders in a cross manner are introduced to decode the spatial-temporal contextual information.
Finally, a Group Decoder is applied to augment the group representation.
We now explain the proposed model in detail.

\textbf{Encoders:} 
Despite being localized for each individual, the representation still lacks emphasis on semantic spatial and temporal context, which play an important role in video analysis.
Thus two parallel encoders are deployed to embed contextual features.
In one branch, we adopt a transformer-based spatial encoder to learn the individual contextual information.
Given the input individual representation $\mathbf{X_I} \in \mathbb{R}^{T \times N \times D}$, we view the temporal dimension as the batch dimension and apply an encoder to exploit spatial context for all frames.
The process of embedding spatial context for $t$-frame can be formulated as:
\begin{align}
&{\mathbf{Q}^{(t)}}=\mathbf{X_I^{(t)}} {W_{tq}}, {\mathbf{K}^{(t)}}=\mathbf{X_I^{(t)}}W_{tk},
{\mathbf{V}^{(t)}}=\mathbf{X_I^{(t)}}W_{tv}\\
&{\mathbf{V}'^{(t)}}=\mathrm{softmax}(\frac{\mathbf{Q}^{(t)}  {\mathbf{K}^{(t)}}^T}{\sqrt{D}})\mathbf{V}^{(t)} + \mathbf{V}^{(t)}\\
&{\mathbf{V}''^{(t)}}=\mathrm{FFN}({\mathbf{V}'^{(t)}})
\end{align}
where $W_{tq}, W_{tk}, W_{tv}$ are learnable parameters shaped as $D \times D$.
$\mathbf{X_{I}^{(t)}}$ denote the individual feature map in $t$-th frame.
$\mathrm{FFN}$ is the feed-forward network in canonical Transformer.
The feature map of all time steps $\{\mathbf{V''}^{(t)}|t=1,...,T\}$ are packed together into $\mathbf{V_s} \in \mathbb{R}^{T \times N \times D}$.

The other parallel temporal encoder is applied to augment the input features with temporal dynamical evolution clues and enrich the temporal context by highlighting the informative features along time dimension for each individual.
The temporal encoder follows the operation of the spatial encoder.
The difference with the above spatial encoder is that the temporal encoder views the spatial dimension as a batch dimension.
We denote the output temporal dynamic features for $n$-th individual as $\mathbf{V''}^{(n)} \in \mathbb{R}^{T \times D}$, and pack the generated feature maps $\{ \mathbf{V}''^{(n)} | n=1,2,...,N\} $ together, represented by $\mathbf{V_t} \in \mathbb{R}^{N \times T \times D}$. 

\textbf{Individual Decoders:}
Individual Decoders are deployed to consider the spatial and temporal contextual information integrally.
Individual Decoders following the standard architecture of Transformer are applied by cross scheme to complementary exploit spatial-temporal context.

For the spatial decoder, $\mathbf{V_s}$ is viewed as the \textit{actor query}, this $N$ individual queries of each frame are transformed into an output embedding by decoder, where the temporal individual embedding $\mathbf{V_t}$ is regarded as \textit{key} and \textit{value}.
The \textit{actor query} captures temporal dynamics from \textit{key} and outputs the updated contextual features.
Simultaneously, we also adopt the other temporal decoder by cross scheme.
Specifically, the spatial embedding $ \mathbf{V_s} $ transposes the time dimension with spatial dimension and can be regarded as the \textit{key} and \textit{value} used by the decoder.
The decoder views temporal context $\mathbf{V_t}$ as \textit{time query} and takes it as input and performs feature embedding.
During the process, the \textit{key} represents spatial features $\mathbf{V_s}$ along with time order and \textit{time query} of each individual looks up the frames of interest in the video.
Finally, the output embedding of these two cross decoders are fused to generate the enhanced individual representation $\mathbf{X_{I}}$.
The concept of these two decoders with cross manner is to exploit semantic associations based on the spatial context and temporal context for augmenting individual representation.

\textbf{Group Decoder:}
Summarizing individuals' interactions in a multi-person scenario is critical to group activity inferring.
We introduce a decoder to augment group representation through individual representation.
The Group Decoder also follows the pure transformer-based decoder.
The difference with the original transformer is that our Group Decoder only contains multi-head cross-attention mechanism and a feed-forward network. 
It takes the enhanced individual representation $\mathbf{X_{I}}$ and group representation $\mathbf{X_G}$ as input.
Motivated by the learned object query proposed by \cite{DETR2020}, we adopt group representation, termed as \textit{group query}, to perform group activity context augmenting from individual representation termed as \textit{key}.
Thus, the \textit{group query} summarize the overall context from augmented individual representation, and group activity prediction is realized by the updated \textit{group query}.
In practice, the output of our Spatial-Temporal Transformer (STT), the enhanced group representation and individual representation, can be used as input of the latter block.
We can stack the STT block repeatedly and learn the underlying semantic representations effectively.
Ablation studies are conducted in Section \ref{sec:ablation} to assess the effectiveness of stacked architecture manner.
\subsubsection{Clustered Attention Mechanism} \label{sec:CAM}
Although Spatial-Temporal Transformer (STT) based on fully-connected attention mechanism is capable of modeling individuals' relations, it contains many irrelevant relations. 
To focus on the crucial group relations, we replace the fully-connected attention with a clustered attention block, called Clustered Spatial-Temporal Transformer (CSTT).
It can group the individuals and exploit intra- and inter-group relations for capturing the global activity context.

We first group the individuals into $\mathbf{C}$ clusters and then compute two types of attentions:
(1) \textbf{intra-group attention} as only \textit{query} and \textit{key} from the same cluster are considered.
(2) \textbf{inter-group attention} as pairwise weighted connection of the clusters are considered.
In detail, we define a set of centroid vectors as $\mathbf{M}=(m_1, ...,m_C) \in \mathbb{R}^{C \times D}$.
We utilize mini-batch k-means clustering algorithm to group \textit{query} into $\mathbf{C}$ clusters adaptively and update the centroid vectors followed the implementation in \cite{convergence1995}.
Our intra-group attention is to refine the action-aware feature of each individual by aggregating the information of relative individuals in the same cluster.
The inter-group attention is to fully model the relations among clusters for facilitating the group activity-aware representation learning.
In detail, the intuitive tactic of building the relations of $\mathbf{C}$ clusters is to regard the clusters' centroid vectors as cluster holistic features $\mathbf{M} \in \mathbb{R}^{C \times D}$.
We first map the cluster holistic feature into \textit{query}, \textit{key} and \textit{value} using linear functions. Then the inter-attention is obtained by dot production followed row-wise $\mathrm{softmax}$.
The cluster features can be calculated by the weight sum of \textit{value}.
The updated centroid vector for each cluster can be broadcast to the actors/individuals belonging to the same cluster.

\subsection{Network Optimization}
Our network is trained in an end-to-end fashion.
In our framework, we directly generate the group activity scores $\bar{y}_{g}$ by group representation obtained from the CSTT.
Similarly, another classiﬁer is adopted to predict the individual's action scores $\bar{y}_{a}$ using individual representation generated by CSTT.
For both tasks, we choose the cross-entropy loss to guide the optimizing process:
\begin{equation}
\mathcal{L}=\mathcal{L}_1(y_g,\bar{y}_{g}) + \lambda \mathcal{L}_2(y_a,\bar{y}_a)
\end{equation}
where $\mathcal{L}_1$ and $\mathcal{L}_2$ denote the cross entropy loss.
$\bar{y}_g$ and $\bar{y}_a$ are the group activities scores and individual actions scores while $y_g$ and $y_a$ represent the ground truth labels for the target group activity and individual actions.
$\lambda$ is the hyper-parameter to balance the two terms. \label{sec:optim}
\section{Experiments and Analysis}
In this section, we experimentally evaluate our proposed network on two widely used datasets.
We first introduce these two available group activity datasets, the Volleyball dataset \cite{hierarchical} and the Collective dataset \cite{cad2009} in Section \ref{sec:dataset}.
Then we describe the training details and parameter setting in Section \ref{sec:implement}.
In section \ref{sec:compare}, we compare our approach with the state-of-the-art.
Finally, a number of ablation studies are conducted to validate the effectiveness of each part within proposed network in section \ref{sec:ablation}.

\subsection{Dataset}\label{sec:dataset}
\noindent
\textbf{Volleyball Dataset.} This dataset \cite{hierarchical} contains 55 volleyball videos with 4,830 labeled frames (3493/1337 for training/testing).
Each clip is annotated with 8 group activity categories: right set, right spike, right pass, right win-point, left set, left spike, left pass, left win-point.
Moreover, the centered frame in each clip is annotated with 9 individual action labels: waiting, setting, digging, falling, spiking, blocking, jumping, moving and standing.\\
\textbf{Collective Activity Dataset.}
This dataset \cite{cad2009} contains 2481 activity clips of 44 video sequences captured by hand-held cameras in the street and indoor scenes.
Group activities classes are annotated with crossing, waiting, queuing, walking and talking.
And the centered frame of each clip are labeled with the bounding boxes of individuals and their individual action classes: NA, crossing, waiting, queuing, walking and talking.
The group activity label is assigned as the largest number of the individual actions in the scenes.
We follow the same dataset splits as previous works \cite{stagnet,actorTR2020}.

\subsection{Implementation details}\label{sec:implement}
For feature extractor, we adopt the Kinetics \cite{Kay:2017:Kinetics} pretrained I3D \cite{2017:I3D} as backbone, separately select 3 frames before and after the annotated frame to be training and testing video clip on both two datasets.
An $1 \times 1$ convolution reduces the channel dimension of $\mathbf{X}_g$ and $\mathbf{X}_d$ to $D=256$.
The RoIAlign with crop size $7 \times 7$ is applied for extracting individuals' feature map with 256-dimension using the ground truth bounding boxes provided by \cite{Social-scene}.
For Volleyball dataset, we resize each frame to $720 \times 1280$ resolution, for the Collective to $480 \times 720$.
For CSTT, we use $1$ encoder/decoder layer with 8 attention heads for all encoders and decoders and set dropout probability to $0.1$.
We choose the number of cluster $C=4$.
Our CSTT is stacked for $\mathbf{b}=3$ blocks.
For the Volleyball dataset we use a batch size of 16 samples and for the Collective dataset we use a batch size of 8 samples.
For both datasets, we adopt ADAM \cite{adam2014} to learn the network parameters.
Initially, we set the learning rate to 0.0001 and decrease by a factor of 10 after 50 and 100 epochs. 
We set the weight term $\lambda=1$. 
Our experiments are all conducted on 8 V100 GPUs.

\begin{table}[t]
\centering
\ttsize
\begin{tabular}{lcccc}
\toprule
Method & Flow &Backbone& \tabincell{c}{Group\\Activity} & \tabincell{c}{Individual\\Action}\\
\midrule
HDTM\cite{hierarchical} && AlexNet  & 81.9 & -\\
SBGAR\cite{SBGAR}      &\checkmark &Inception-v3&67.6& -\\
CERN\cite{CERN}         && VGG16    &  83.3& 69.1 \\
stagNet\cite{stagnet2018}&&VGG16   & 89.3 & -\\
HRN \cite{hier2018}&  &VGG19     & 89.5 & -\\
SSU \cite{Social-scene}&&Inception-v3&90.6&81.8\\
ARG \cite{ARG2019}&    &Inception-v3&92.5&83.0\\
CRM \cite{CRM2019}   &\checkmark  &   I3D    &93.0   &-\\
Gavrilyuk\etal\cite{actorTR2020}&\checkmark&I3D&93.0&83.7\\
Gavrilyuk\etal\cite{actorTR2020}&\checkmark&I3D+HRnet&94.4&\textbf{85.9}\\
Ehsanpour \etal \cite{joint2020}&\checkmark&I3D&93.1&83.3\\
Pramono \etal \cite{empowering2020}&\checkmark&I3D&94.1&81.9\\
Pramono \etal \cite{empowering2020}&\checkmark&I3D+Pose+FPN&95.0&83.1\\

\hline
Ours \textrm{w/o} GRG && Inception-v3 &93.4 &83.2\\
Ours & &Inception-v3 &94.1&83.7\\
Ours &\checkmark &I3D          &94.9&84.0\\
Ours &\checkmark &I3D+Pose     &\textbf{95.7}& 85.6\\
\bottomrule
\end{tabular}
\caption{Comparisons with the state-of-the-art methods on Volleyball dataset in terms of Acc.\%. ``Flow'' denotes additional optical flow input.}
\label{tab:volley}
\end{table}

\subsection{Comparison with the State of the Art} \label{sec:compare}
We compare our approach with the state-of-the-art methods in two widely adopted datasets, the Volleyball dataset and the Collective dataset separately.
For a fair comparison with previous methods, we not only show the results with Pose feature and I3D backbone with RGB and optical flow \cite{flow} features but also report the results with Inception-v3 only using RGB features.\\
\textbf{Volleyball dataset.}
The results are listed in Table \ref{tab:volley}, our method outperforms all of the aforementioned methods with a considerable margin for activity accuracy.
To show the effectiveness of our CSTT clearly, we report the experimental result of the proposed model without GRG, which discards GRG and introduces a learned query to be the initialized group representation.
It is noting that our model with only RGB features outperforms many previous works \cite{joint2020,actorTR2020,empowering2020} though they utilize multi-steps backbone (optical flow, Pose, FPN, Spatial Position), since our GroupFormer exploits both spatial and temporal dependencies integrally.
Furthermore, we take pose information as additional input and achieve the best results 95.7\% comparing with the previous methods \cite{actorTR2020,empowering2020}.It's because that we model the spatial and temporal context in a joint model and augment individual and group representation in parallel.
Specifically, compared with \cite{actorTR2020}, the performance of our transformer based method is boosted up by a significant 1.3\%, up to 95.7\%, indicating that our CSTT can predict the spatio-temporal dynamic of interactions and better enrich group context.  \\
\textbf{Collective dataset.}
We further provide detail comparisons with previous methods listed in Table \ref{tab:cad} on the Collective dataset.
The results with Inception-v3 backbone using RGB features can reach 93.6\%, which  outperforms most previous methods that use additional optical flow features.
Considering the accuracy of 96.3\%, a considerable improvement for our model with I3D+Pose backbones is achieved compared to the previous methods \cite{empowering2020,actorTR2020,joint2020}.
It is worth noting that \cite{actorTR2020} achieves 92.8\% which is 1.9\% lower than our result, showing  the benefits of spatial-temporal transformer over their spatial transformer based approach.
\subsection{Ablation Studies}\label{sec:ablation}
To validate the effectiveness of the different parts of GroupFormer, we perform ablation studies on the validation set of Volleyball dataset and use group activity accuracy and individual action accuracy as our evaluation metrics.
To dispel any concerns that the improvement is simply from additional optical flow and pose information, we only utilize the ImageNet \cite{imagenet2012} pretrained Inception-v3 \cite{inception2016} as our backbone to extract features from RGB clips in our ablation studies.

\begin{table}[t]
\centering
\ttsize
\begin{tabular}{lcccc}
\toprule
Method & Flow &Backbone& \tabincell{c}{Group\\Activity}\\
\midrule
HDTM\cite{hierarchical} && AlexNet  & 81.5\\
CERN\cite{CERN}         && VGG16    & 87.2 \\
stagNet\cite{stagnet2018}&&VGG16   & 87.7\\
ARG \cite{ARG2019}&    &Inception-v3&91.0\\
CRM \cite{CRM2019}   &\checkmark  &   I3D    &85.8\\
Gavrilyuk\etal\cite{actorTR2020}&\checkmark&I3D&92.8\\
Ehsanpour \etal \cite{joint2020}&\checkmark&I3D&89.4\\
Pramono \etal \cite{empowering2020}&\checkmark&I3D&93.9\\
Pramono \etal \cite{empowering2020}&\checkmark&I3D+Pose+FPN&95.2\\

\hline
Ours & &Inception-v3 &93.6&\\
Ours &\checkmark &I3D          &94.7&\\
Ours &\checkmark &I3D+Pose     &\textbf{96.3}&\\
\bottomrule
\end{tabular}
\caption{Comparison with the state-of-the-art methods on Collective dataset in terms of Acc.\%. ``Flow'' denotes additional optical flow input.}
\label{tab:cad}
\end{table}

\begin{table}[t]
\centering
\ttsize
\begin{tabular}{lcc}
\toprule
Manner & \tabincell{c}{Group Activity} & \tabincell{c}{Individual Action}\\
\midrule
Baseline    &91.0& 82.1\\
Spatial only &     91.8          &     82.2               \\
Stacked&            92.6           &       82.8      \\
Parallel &           92.2      &   82.9    \\
\textbf{Ours}       &     \textbf{94.1}    &     \textbf{83.7}        \\
\bottomrule
\end{tabular}
\\
\caption{{Ablation study on different variants architectures.}}

\label{tab:variant}
\end{table}

\noindent
\textbf{Variants of Spatial-Temporal Relations Modeling.}
To measure the importance of the contextual information gathered by our CSTT, we conduct ablation studies with the following variants.
(1) baseline: we replace CSTT with a FC layer and a Group Decoder is followed.
(2) \textit{spatial manner}: this variant involves a spatial encoder followed by a Group Decoder.
(3) \textit{stacked manner}: this variant contains stacked spatial and temporal encoders, and a Group Decoder.
(4) \textit{parallel manner}: this variant consists of parallel encoders, which deal with spatial and temporal context separately and aggregates them by `sum', and then it follows a Group Decoder.
Apart from the manner of architecture, the other settings are all the same.
Table \ref{tab:variant} shows that adopting spatial and temporal in parallel manner improves the performance from 91.0\% to 92.2\%.
Meanwhile, the stacked manner performs slightly better than the models in parallel manner.
And our CSTT using Individual Decoders, which build a bridge for jointly exploiting the spatial-temporal contextual information in a cross manner, makes a remarkable performance boost by 3.1\% comparing to the baseline.
It stresses that learning the spatial-temporal context integrally is effective and important for group activity recognition. 
\begin{table}[t]
\centering
\ttsize
\begin{tabular}{ccc}
\toprule
Clusters & \tabincell{c}{Group Activity} & \tabincell{c}{Individual Action}\\
\midrule
1(w/o cluster) &      93.4                   &    83.1                      \\
2&       93.7                 &    83.6          \\
3 &      93.8           &    \textbf{83.8}    \\
4 &      \textbf{94.1}               &{83.7}\\
6&          93.4& 83.5 \\
\bottomrule
\end{tabular}
\caption{Comparisons of different clusters choices on Volleyball dataset. Cluster set to $1$ demonstrate that we adopt original Spatial-Temporal Transformer.}
\label{tab:cluster}
\end{table}

\begin{table}[t]
\centering
\ttsize
\begin{tabular}{cccc}
\toprule
{Intra-Attn}&{Inter-Attn}& \tabincell{c}{Group \\Activity}& \tabincell{c}{Individual\\ Action} \\
\midrule
 &    & 93.4 &83.1\\
\checkmark &  &93.8&83.4\\
    & \checkmark &93.6&83.5\\
\checkmark&\checkmark &\textbf{94.1}&\textbf{83.7}\\

\bottomrule
\end{tabular}
\caption{Comparisons of different cluster attention combinations. Intra-Attn and Inter-Attn denote intra-group attention and inter-group attention respectively.}
\label{tab:inter_intra}
\end{table}

\begin{table}[t]
\centering
\ttsize
\begin{tabular}{ccc}
\toprule
Blocks & \tabincell{c}{Group Activity} & \tabincell{c}{Individual Action}\\
\midrule
0 & 91.0 &82.1\\
1 &      93.6                   &    83.4                      \\
2&       93.8                 &    83.7          \\
3 &      \textbf{94.1}           &    \textbf{83.7}    \\
4 &      93.9               & 83.6\\
\bottomrule
\end{tabular}
\caption{Comparisons of different setting choices for the number of CSTT blocks.}
\label{tab:blocks}
\end{table}

\begin{figure*}[t]
\centering
\includegraphics[width=0.9\linewidth]{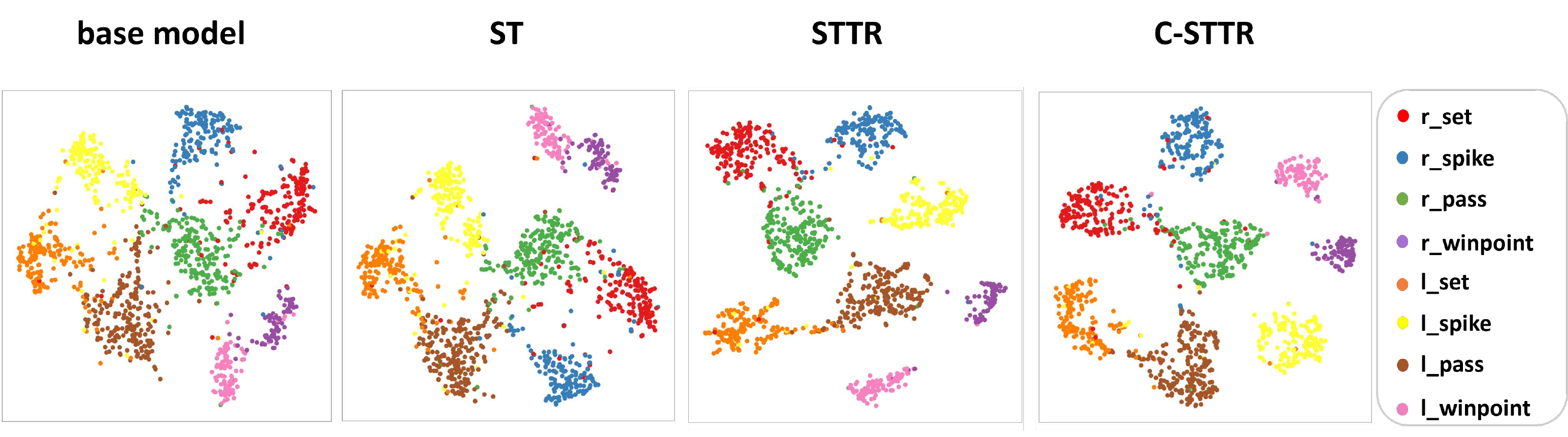}
\caption{Feature embedding visualizations of the validation set of Volleyball dataset using t-SNE \cite{tsne2008} by different model variants.
Each clip is visualized as a point and clips belonging to the same group activity have the same color. Best viewed in color.}
\label{fig:visulization}
\end{figure*}

\begin{figure*}[t]
\centering
\includegraphics[width=0.9\linewidth]{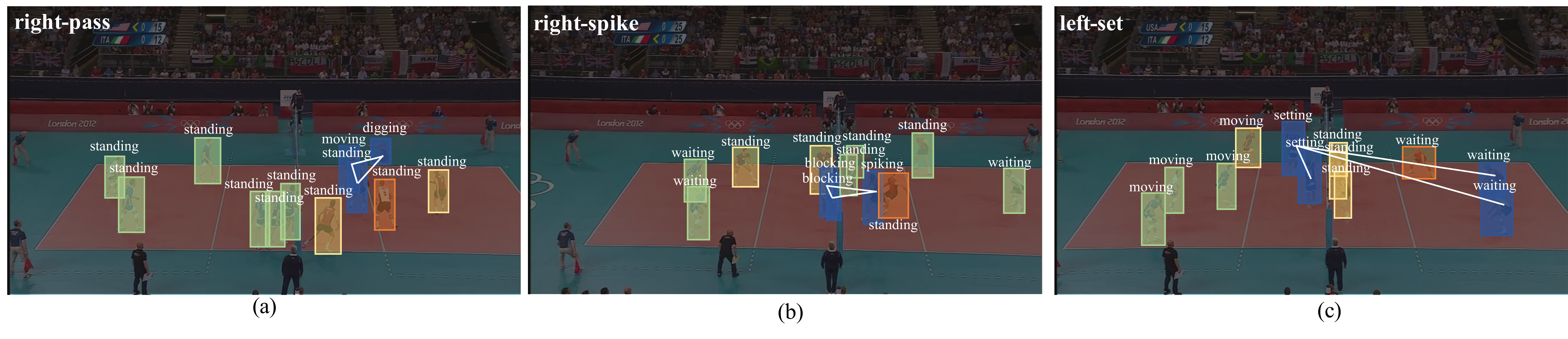}
\caption{
Visualization of some representative clustered individuals. Each video scenario contains the groundtruth labels of activity and individuals.
The color of each individual represents the clusters that it belongs to. 
The white lines represent the close relations in the cluster.}
\label{fig:visulization2}
\end{figure*}

\noindent
\textbf{Clustered Attention Manner.}
Here we investigate the performances of different cluster choices.
The results on Volleyball dataset are displayed in Table \ref{tab:cluster}.
The STT without clustered attention mechanism can achieve 93.4\%, indicating that our proposed model (Spatial-Temporal Transformer) is already capable of learning rich spatial-temporal context for group activity recognition. 
Meanwhile, we replace the fully-connected attention with clustered attention mechanism called CSTT, leading to a significant boost from 93.4\% to 94.1\%.
It indicates the effectiveness of our clustered attention mechanism.
In addition, we gradually modify the number of clusters and find that the number of cluster set to $4$ can reach the best result.
Finally, we also evaluate the effectiveness of our two intra-group and inter-group attention. Table \ref{tab:inter_intra} reports the detailed results. 
As expected, we can find that these two attention manner can always contribute to certain performance boosts and the combinations of them reach the best performance.

\noindent
\textbf{Investigation of Number of CSTT Blocks.}
Our clustered Spatial-Transformer can be stacked several blocks to enhance spatial-temporal information.
It is essential to evaluate the influence of block number setting.
As shown in Table \ref{tab:blocks}, the first column of the table lists the corresponding number of our Clustered Spatial-Transformer.
When the number of blocks is set to $0$,  a FC layer is adopted to replace our CSTT to embed features.
We find that a single block outperform baseline by 2.6\%, which demonstrate the effectiveness of CSTT.
Stacked $3$ CSTT blocks reach best results while performance slightly degrades at using $4$ blocks.


\subsection{Visualization}
\noindent
\textbf{Feature Embedding Visualization.}
Figure \ref{fig:visulization} displays the t-SNE \cite{tsne2008} visualization of the video representation learned by our model variants:
(1) {baseline}: replace CSTT with simple FC layers.
(2) \textit{Stacked} ST: using stacked spatial temporal transformers.
(3) {STT}: Spatial-Temporal Transformer without Clustered attention mechanism.
(4) {CSTT}: our Clustered Spatial-Temporal Transformer.
In detail, we feed our video representation of the validation set on the Volleyball dataset into 2-dimensional map using t-SNE.
We can find that adopting our STT has performed quite well compared with \textit{Stacked} ST.
It is noting that our CSTT makes separation better and reaches best results.
These visualization results demonstrate that our model is conducive to recognize group activity.

\noindent
\textbf{Cluster Visualization.}
We visualize several examples of the clustered individuals in Figure \ref{fig:visulization2}.
With clusterd attention mechanism, individuals are grouped into sets of nodes, and each set of nodes is closely connected internally. Through exploiting intra- and inter-group relations, it's easier for the model to capture crucial interaction information and learn activity-aware semantic representations.

\section{Conclusion}
We propose a new transformer-based architecture termed GroupFormer, which models the spatial-temporal contextual representation for inferring group activity.
Furthermore, we introduce a cluster attention mechanism to group the individuals and exploit intra- and inter-group relations together for better group informative features.
We perform extensive experiments on two benchmarks.
The results show that our GroupFormer outperforms most state-of-the-art methods by a considerable margin.
{\small
\bibliographystyle{ieee_fullname}
\bibliography{article}
}

\end{document}